\documentclass[10pt,twocolumn,letterpaper]{article}

\usepackage{cvpr}
\usepackage{multirow}
\usepackage{graphicx}
\usepackage{amsmath}
\usepackage{amssymb}
\usepackage{booktabs}
\usepackage[table]{xcolor}
\usepackage{makecell}
\usepackage[misc]{ifsym}

\makeatletter
\renewcommand*{\@fnsymbol}[1]{\ensuremath{\ifcase#1\or \dagger\or \ddagger\or
   \mathsection\or \mathparagraph\or \|\or **\or \dagger\dagger
   \or \ddagger\ddagger \else\@ctrerr\fi}}
\makeatother

\renewcommand\toprule{\specialrule{1.5pt}{1pt}{0pt}}
\renewcommand\midrule{\specialrule{0.5pt}{0.1pt}{0.1pt}}
\renewcommand\bottomrule{\specialrule{1.5pt}{0pt}{1pt}}

\usepackage[pagebackref,breaklinks,colorlinks]{hyperref}

\usepackage[capitalize]{cleveref}
\crefname{section}{Sec.}{Secs.}
\Crefname{section}{Section}{Sections}
\Crefname{table}{Table}{Tables}
\crefname{table}{Tab.}{Tabs.}

\begin{document}

\title{Teach-DETR: Better Training DETR with Teachers}

\author{Linjiang Huang\textsuperscript{\rm 1,2} \quad
        Kaixin Lu \textsuperscript{\rm 2,3} \thanks{This work is done during a research assistant at CPII.} \quad
        Guanglu Song \textsuperscript{\rm 4} \quad
        Liang Wang \textsuperscript{\rm 6} \\
        Si Liu \textsuperscript{\rm 5} \quad
        Yu Liu \textsuperscript{\rm 4} \quad
        Hongsheng Li\textsuperscript{\rm 1,2} \thanks{Corresponding author.}\\
\textsuperscript{\rm 1}CUHK-SenseTime Joint Laboratory, The Chinese University of Hong Kong \\
\textsuperscript{\rm 2}Centre for Perceptual and Interactive Intelligence, Hong Kong \\
\textsuperscript{\rm 3}Shanghai University \quad
\textsuperscript{\rm 4}Sensetime Research \quad
\textsuperscript{\rm 5}Beihang University \\
\textsuperscript{\rm 6}Institute of Automation, Chinese Academy of Sciences \\
{\tt\small ljhuang524@gmail.com, hsli@ee.cuhk.edu.hk}
}

\maketitle

\begin{abstract}
In this paper, we present a novel training scheme, namely Teach-DETR, to learn better DETR-based detectors from versatile teacher detectors. We show that the predicted boxes from teacher detectors are effective medium to transfer knowledge of teacher detectors, which could be either RCNN-based or DETR-based detectors, to train a more accurate and robust DETR model. 
This new training scheme can easily incorporate the predicted boxes from multiple teacher detectors, each of which provides parallel supervisions to the student DETR.
Our strategy introduces no additional parameters and adds negligible computational cost to the original detector during training. During inference, Teach-DETR brings zero additional overhead and maintains the merit of requiring no non-maximum suppression. Extensive experiments show that our method leads to consistent improvement for various DETR-based detectors. Specifically, we improve the state-of-the-art detector DINO \cite{zhang2022dino} with Swin-Large \cite{liu2021swin} backbone, 4  scales of feature maps and 36-epoch training schedule, from 57.8\% to 58.9\% in terms of mean average precision on MSCOCO 2017 validation set. Code will be available at \url{https://github.com/LeonHLJ/Teach-DETR}.
\end{abstract}

\section{Introduction} \label{sec:intro}

Object detection is a fundamental task in computer vision, which aims to localize the boxes and categories of objects of interest. Due to the significant successes in various fields, deep learning has been the prevailing solution for object detection. The previous development is rapidly advanced by a series of RCNN-based methods, \eg, Faster-RCNN \cite{ren2015faster}, YOLO \cite{redmon2016you}, RetinaNet \cite{lin2017focal}, \etc. 
\begin{table}[!t]
\begin{center}
\caption{Detection results of utilizing existing knowledge distillation methods and our Teach-DETR for object detection. The student DETR is an $\mathcal{H}$-Deformable-DETR \cite{jia2022detrs} with Swin-small backbone, 300 queries and 12-epoch training schedule. The teacher detector is the Mask-RCNN \cite{he2017mask} with Swin-small backbone and 36-epoch training schedule.}
\vspace{-1mm}
\label{table:knowledge_distill}
\begin{tabular}{l|cccc}
\toprule
\rowcolor{black!20}
\textbf{Method} & \textbf{AP} & \textbf{AP$_{50}$} & \textbf{AP$_{75}$} \\
\hline
\hline
Student ($\mathcal{H}$-Def-DETR) & 52.5 & 71.1 & 57.3 \\
Teacher (Mask-RCNN) & 46.4 & 67.0 & 50.5 \\
\midrule
FKD \cite{zhang2020improve} & 52.0 ($\downarrow0.5$) & 70.5 & 57.1 \\
DeFeat \cite{guo2021distilling} & 52.2 ($\downarrow0.3$) & 70.8 & 57.1 \\
FGD \cite{yang2022focal} & 51.7 ($\downarrow0.8$) & 70.0 & 57.0 \\
\midrule
Teach-DETR & 53.5 ($\uparrow1.0$) & 72.0 & 58.5 \\
\bottomrule
\end{tabular}
\vspace{-3mm}
\end{center}
\end{table}

Recently, DEtection TRansformer (DETR) \cite{carion2020end} introduces the Transformer \cite{vaswani2017attention} to serve as detection heads and achieves impressive performance without the need of non-maximum suppression (NMS). The design of DETR is very different from the RCNN-based detectors \cite{ren2015faster,redmon2016you,lin2017focal}. On top of the feature map from a backbone network, a Transformer encoder-decoder regards and assigns object detection as a set prediction problem \cite{kuhn1955hungarian}. Nevertheless, through the rapid development of classical detectors in the past decade, there have been many mature designs and tens of well-learned RCNN-based detectors with promising performance.
We would like to ask the following question: given well-trained RCNN-based detectors and DETR-based detectors, can we effectively transfer their knowledge to train a more accurate and robust DETR model?
Knowledge distillation (KD) \cite{hinton2015distilling} seems to be a plausible way to transfer knowledge from teacher detectors to DETRs. However, our experiments in Tab. \ref{table:knowledge_distill} show that, existing knowledge distillation methods \cite{zhang2020improve,guo2021distilling,yang2022focal} cannot effectively transfer knowledge from RCNN-based detectors to DETR-based detectors. Native KD methods sometimes even result in worse performance to student DETR-based detectors.

In this paper, we propose a novel training scheme, namely Teach-DETR, to construct proper supervision from teacher detectors for training more accurate and robust DETR models. We find that the predicted bounding boxes of teacher detectors can serve as an effective bridge to transfer knowledge between RCNN-based detectors and DETR-based detectors. We make the following important observations: first, the bounding boxes are the unbiased representation of objects for all detectors, which would not be affected by the discrepancies in detector frameworks, label assignment strategies, \etc. Second, introducing more box annotations would greatly unleash the capacity of object queries by training them with various one-to-one assignment patterns \cite{jia2022detrs,chen2022group}, and thus improve training efficiency and detection performance. Finally, leveraging the predicted bounding boxes of teacher detectors would not introduce additional architectures, since it has the similar format to the ground truth (GT) boxes.
Nevertheless, it is non-trivial to integrate the GT annotations with the auxiliary supervisions. Due to the one-to-one matching of query-label assignment of DETR-based detectors, incorporating the auxiliary supervisions would be harmful to maintain the key signature of DETR of enabling inference without NMS.
To tackle the challenge of transferring knowledge across different types of detectors, we propose a solution that aligns with our observations and utilize output boxes and scores of the teacher detectors as extra \emph{independent} and \emph{parallel} supervisions for training the queries' outputs.
In addition, we employ the output scores of teacher detectors to measure the quality of the auxiliary supervisions and accordingly assign the predicted boxes different importance for weighting their losses. The ambiguity between GT boxes and the imperfect boxes from teachers can be alleviated and therefore enhance the importance of GT boxes.

Our Teach-DETR is a versatile training scheme and can be integrated with various popular DETR-based detectors without modifying their original architectures. Moreover, our framework has no requirement on teacher architectures, it can be RCNN-based detectors or DETR-based detectors, which is more general to various types of teachers.
During training, our method introduces no additional parameters and adds negligible computational cost upon the original detector. During inference, our method brings zero additional overhead. We conduct extensive experiments to prove that Teach-DETR can consistently improve the performance of DETR-based detectors. For example, our approach improves the state-of-the-art detector DINO \cite{zhang2022dino} with Swin-Large \cite{liu2021swin} backbone, 4 scales of feature maps and 36-epoch training schedule, from 57.8\% to 58.9\% in terms of mean average precision on MSCOCO 2017 \cite{lin2014microsoft} validation set, demonstrating effectiveness of our method even when applied to a state-of-the-art high-performance detector.

\section{Related Works}

\subsection{DETR and Its Variants}

DEtection TRansformer (DETR) \cite{carion2020end} first applies Transformers \cite{vaswani2017attention} to object detection and achieves impressive performance without the requirement for non-maximum suppression. Despite its significant progress, the slow convergence of DETR is a critical issue which hinders its practical implementation. There are many researchers devote to address the limitations and achieve promising speedups and better detection performance. Some works focus on designing better Transformer layers. For example, Deformable-DETR \cite{zhu2021deformable} introduces the multi-scale deformable attention scheme, which sparsely sampling a small set of key points referring to the reference points. 
Many works \cite{gao2021fast,meng2021conditional,wang2022anchor,liu2022dab} argue that the slow convergence of DETR mainly lies in that DETR could not rapidly focus on regions of interest. To address this issue, SMCA \cite{gao2021fast} and DAB-DETR \cite{liu2022dab} propose to modulate the cross-attentions of DETR, making queries to attend to restricted local regions. Anchor-DETR \cite{wang2022anchor} retrospects the classical anchor-based methods and propose to use anchor points as object queries, one of which response for a restricted region. Conditional-DETR \cite{meng2021conditional} decouples each query into a content query and a positional query, which has a clear spatial meaning to limit the spatial range for the content queries to focus on the nearby region. Some recent approaches \cite{li2022dn,zhang2022dino} ascribe the slow convergence issue to the unstable bipartite matching \cite{kuhn1955hungarian}. They introduce an auxiliary query denoising task to stabilize bipartite graph matching and accelerate model convergence. Two recent approaches \cite{jia2022detrs,chen2022group} look at this problem from a new perspective, arguing that slow convergence results from one-to-one matching. They propose to duplicate the GT boxes to support one-to-many matching in DETR. Unlike previous approaches, we propose a new training scheme to learn better DETR-based detectors from teachers. Theoretically, our approach is orthogonal to previous methods, and thus can further improve their performance.

\subsection{Transfer Knowledge for Object Detection}

For object detection, the commonly used way of leveraging teacher detectors is knowledge distillation (KD) \cite{hinton2015distilling}.
KD is usually applied to decrease the model complexity while improving the performance of the smaller student model.
Compared to KD in classification, KD in object detection should consider more complex structure of boxes and somewhat cumbersome pipelines, and there are more hints, \eg, anchor predictions \cite{chen2017learning,sun2020distilling,zheng2022localization}, proposal ranking results \cite{li2022knowledge}, object-related features \cite{li2017mimicking,wang2019distilling,dai2021general,zhixing2021distilling}, contextual features \cite{guo2021distilling,yang2022focal} and relations among features \cite{yao2021g,yang2022focal}, can be used. For DETR, there are few works have used KD. ViDT \cite{song2021vidt} try to utilize KD between output queries of teacher DETR and student DETR to reduce compution cost.
In contrast, our method do not mean to conduct knowledge distillation or model compression. 
The core idea of Teach-DETR is to transfer knowledge of various teachers to train a more accurate and robust DETR-based detector. Therefore, the teacher detectors can be smaller or perform worse than the ``student'' DETR.
Besides, our approach could be a meaningful exploration, providing a way to distill knowledge from any detectors to DETRs for future KD methods.

\section{Method} \label{sec:method}

Our method aims to transfer knowledge of teacher detectors to train a more accurate and robust DETR-based detector. (i) We introduce the predicted bounding boxes of multiple teacher detectors, which can be RCNN-based detectors, DETR-based detectors or even both types of detectors, to serve as auxiliary supervisions for training DETR; (ii) We weight the corresponding losses of the auxiliary supervisions with the predicted box scores from teacher detectors. In this section, we will first retrospect DETR \cite{carion2020end} (Sec. \ref{sec:preliminary}) and present details of Teach-DETR in (Sec. \ref{sec:our_method}).

\subsection{Preliminary}  \label{sec:preliminary}

DEtection TRansformer (DETR) \cite{carion2020end} usually consists of a backbone (\eg, ResNet \cite{he2016deep}, Swin Transformer \cite{liu2021swin}, \etc), and a Transformer encoder-decoder for box classification and regression. The image features are first extracted by the backbone and then forwarded to the Transformer encoder to aggregate global features into each image token. The Transformer decoder takes a fixed-size set of $N$ queries and perform self-attention among object queries and cross-attention between queries and image tokens. After queries are processed by the Transformer decoder,  they are fed into the detection heads to obtain $N$ box predictions $\mathbf{P}=\left\{\mathbf{p}_0, \mathbf{p}_1, \dots, \mathbf{p}_n\right\}$. DETR employs the Hungarian matching \cite{kuhn1955hungarian} to establish bipartite matching $\hat{\sigma}$ between the ground truth boxes and the predictions with the lowest cost:
\begin{equation} \label{eq:hungarian}
    \hat{\sigma}=\underset{\sigma \in \mathfrak{S}_N}{\arg \min } \sum_i^N \mathcal{L}_{\text {match}}\left(y_i, \hat{y}_{\sigma(i)}\right),
\end{equation}
where $\mathfrak{S}_N$ is a permutation of $N$ elements and $\mathcal{L}_{\text {match}}$ is the box classification and regression cost between the ground truth $y_i$ and the prediction $\hat{y}_{\sigma(i)}$ of the query $\sigma(i)$. To achieve the size $N$ for all batches, the GTs are padded with $\varnothing$ (no object).
\begin{figure}[!t]
  \centering
  \includegraphics[width=1.0\columnwidth]{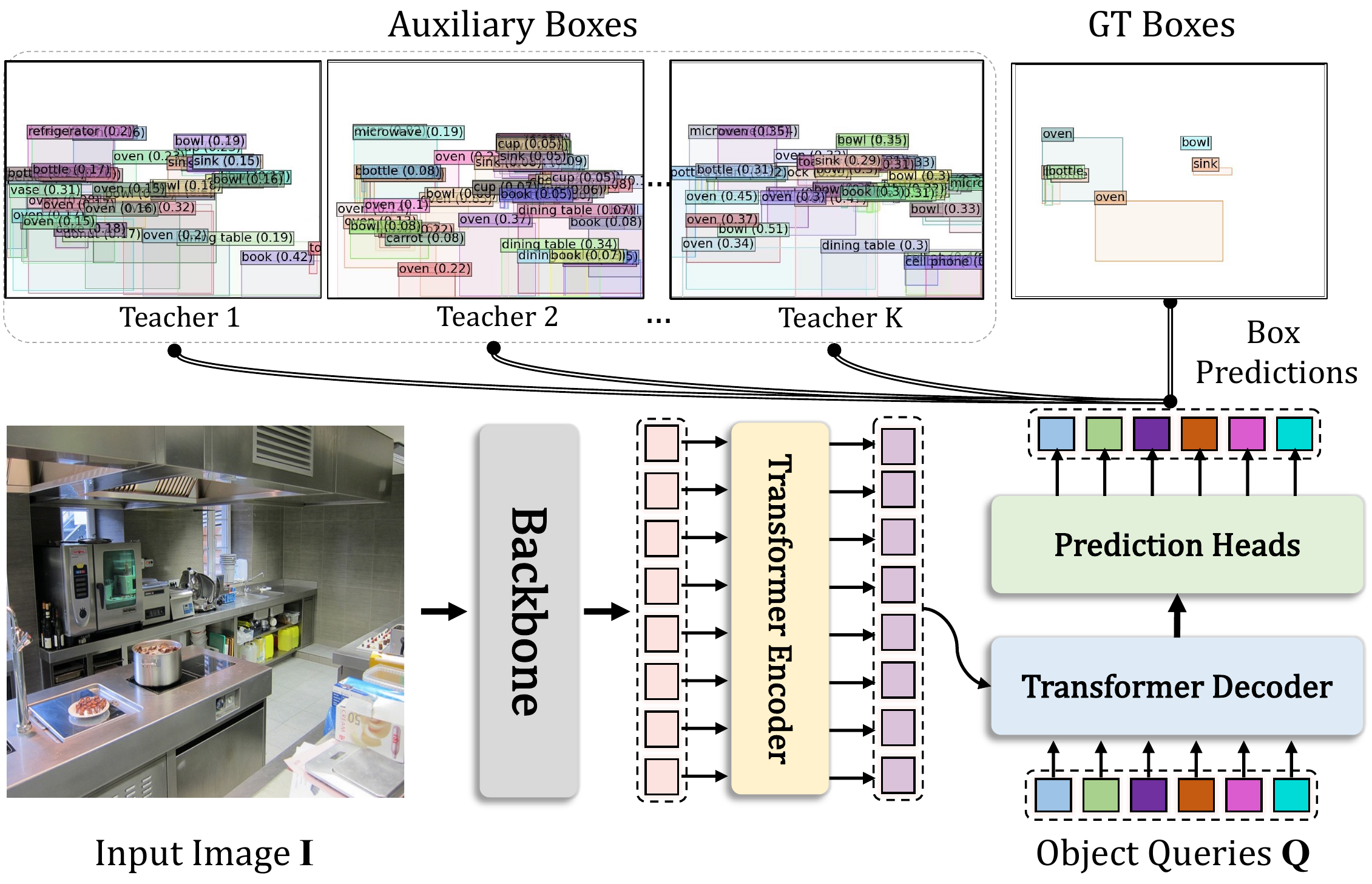}
\caption{The pipeline of our proposed Teach-DETR. The auxiliary supervisions from multiple teacher detectors are used to conduct bipartite matching and weighting independently. The final training loss is the sum of the losses of $K$ teachers and the original losses of the GT boxes.}
\label{fig:pipeline}
\vspace{-1mm}
\end{figure}

\subsection{Training DETR with Teachers} \label{sec:our_method}
\paragraph{Exploration of DETR with knowledge distillation.}
We aim at utilizing various types of teacher detectors to improve the detection performance of DETR-based detectors. A straightforward solution is to utilize knowledge distillation (KD) methods. However, existing KD methods show limited effectiveness when conducting knowledge transfer between RCNN-based detectors and DETR-based detectors. We follow several state-of-the-art KD approaches \cite{zhang2020improve,guo2021distilling,yang2022focal} to perform feature imitation between the features of Mask RCNN's FPN and the multi-scale features generated by Swin-Small backbone. In DeFeat \cite{guo2021distilling}, we also try to distill the classification logits\footnote{Slightly different from the original method, we feed the bounding boxes produced by DETR into Mask RCNN to perform ROI-align and generate the category predictions for distillation.} of the predicted boxes.
As shown in Tab. \ref{table:knowledge_distill}, since there are much differences between the detection pipelines of the RCNN-based and DETR-based detectors, it is very difficult to transfer knowledge from RCNN-based detectors to DETR-based ones. Existing KD methods even results in worse performance to the student detector.

\vspace{-4mm}
\paragraph{Auxiliary supervisions from teacher detectors.} 
In light of the above challenges, we propose to leverage the predicted bounding boxes of teachers as the auxiliary supervisions for better training DETRs. 
There are two main reasons for using the predicted boxes for knowledge transfer.
First, the bounding box is the unbiased representation of results of all detectors, which would not be affected by the discrepancies in different detector frameworks, so it can serve as a good medium to transfer the knowledge of teachers to student DETRs. Second, introducing more box annotations would largely unleash the capacity of object queries by providing more positive supervisions \cite{jia2022detrs,chen2022group}, and thus improve training efficiency and detection performance. The whole pipeline of our Teach-DETR is shown in Fig. \ref{fig:pipeline}. 
\begin{table}[!t]
\begin{center}
\caption{Results on MSCOCO validation set with different combinations of GT boxes and auxiliary boxes as learing targets. The student is the $\mathcal{H}$-Deformable-DETR \cite{jia2022detrs} with Swin-small backbone, 300 queries and 12-epoch training schedule. Auxiliary boxes (if applicable) from the Swin-small Mask-RCNN \cite{he2017mask} are applied to the one-to-one branch \cite{jia2022detrs}. ``Score'' denotes weighting supervisions with teacher's predicted confidence scores.}
\vspace{-1mm}
\label{table:gt_auxilarity_boxes_relations}
\resizebox{1\columnwidth}{!}{
\begin{tabular}{l|cccc}
\toprule
\rowcolor{black!20}
\textbf{Method} & \textbf{AP} & \textbf{AP$_{50}$} & \textbf{AP$_{75}$} \\
\hline
\hline
Baseline & 52.5 & 71.1 & 57.3 \\
\midrule
Concat GT \& Aux & 41.9 ($\downarrow10.6$) & 56.6 & 45.6 \\ 
Parallel GT \& Aux & 53.1 ($\uparrow0.6$) & 71.6 & 58.3 \\
Parallel GT \& Aux w/ Score & 53.5 ($\uparrow1.0$) & 72.0 & 58.2 \\
\bottomrule
\end{tabular}}
\vspace{-3mm}
\end{center}
\end{table}

However, since the one-to-one matching is the critical design of DETRs to discard the NMS during inference, naively introducing the auxiliary supervisions would result in ambiguous training targets. How to balance the contributions between different sets of teachers' output boxes and GT boxes is a problem. In Tab. \ref{table:gt_auxilarity_boxes_relations}, we show that simply concatenating the auxiliary boxes and the GT boxes greatly deteriorates the performance.

We try to address the above issue by conducting the one-to-one matching between GT boxes and object queries, and between each teacher's boxes and object queries independently. The matchings are also properly weighted according to the teachers' confidences on the predicted boxes. Specifically, for an input image, given $K$ teacher detectors, we can obtain $K$ sets of detection boxes, for the $i$-th teacher, it contains for example $M$ predicted bounding boxes $\left\{\bar{y}^i_1, \dots, \bar{y}^i_m\right\}$. Each of these predicted boxes $\bar{y}^i_j$ contains the estimated box size and location, the predicted category, and the predicted confidence score $\bar{s}^i_j$. We apply one-to-one assignment between queries not only to the GT, but also to the auxiliary supervision of each teacher detector independently and thus we can obtain $K$ groups of matching:
\begin{equation}
\begin{gathered}
    \hat{\sigma}_1 =\underset{\hat{\sigma}_1 \in \mathfrak{S}_N}{\arg \min } \sum_i^N \mathcal{L}_{\text {match}}\left(\bar{y}^1_i, \hat{y}_{\sigma_1(i)}\right), \\
    \vdots \\
    \hat{\sigma}_k =\underset{\hat{\sigma}_k \in \mathfrak{S}_N}{\arg \min } \sum_i^N \mathcal{L}_{\text {match}}\left(\bar{y}^k_i, \hat{y}_{\sigma_k(i)}\right), \\
\end{gathered}
\end{equation}
where $\hat{\sigma}_k$ is the matching result of the $k$-th teacher's auxiliary supervisions. To balance the teacher boxes' contributions, we propose to take advantage of the confidence scores of boxes predicted by the teacher detectors. Intuitively, the confidence scores can represent the quality of bounding boxes to some extent. Therefore, we utilize the confidence scores as weights to modulate the corresponding loss of each box's classification and regression loss. The final loss $\mathcal{L}^i_j$ of the auxiliary box $\bar{y}^i_j$ can be formulated as
\begin{equation} \label{eq:loss}
    \mathcal{L}^i_j= \bar{s}^i_j \cdot (\lambda_{\text {cls}} \mathcal{L}^i_{\text {cls}, j} + \lambda_{\text {box}} \mathcal{L}^i_{\text {box}, j}),
\end{equation}
where $\mathcal{L}_{\text {cls}}$ denotes the classification loss and $\mathcal{L}_{\text {box}}$ denotes the box regression loss, including the $\ell_1$ loss and the GIoU loss \cite{carion2020end}. $\lambda_{\text {cls}}$ and $\lambda_{\text {box}}$ are balancing weights. For those negative queries, we assign them with a moderate score of 0.5.
The final training loss is the sum of the losses of $K$ teachers and those of the GT boxes.
\begin{figure}[!t]
  \centering
  \includegraphics[width=0.67\columnwidth]{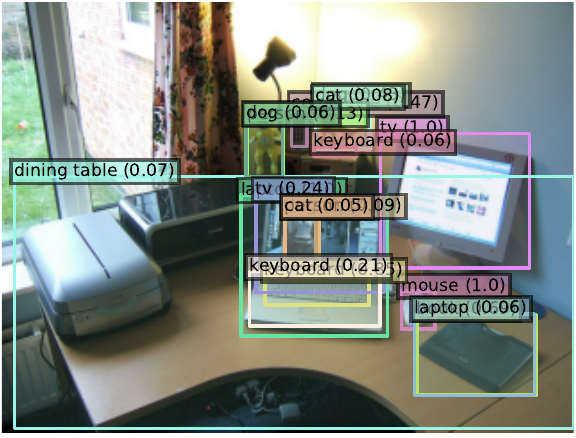}
\caption{Visualization of the boxes detected from a Swin-S Mask RCNN \cite{he2017mask} teacher.}
\label{fig:aux_supervision}
\vspace{-2mm}
\end{figure}

As shown in Tab. \ref{table:gt_auxilarity_boxes_relations}, with the auxiliary boxes, the performance of Swin-Small $\mathcal{H}$-Deformable-DETR \cite{jia2022detrs} increases by 0.6\%. Benefited further from the use of confidence scores, our method can finally improve the AP by 1.0\% on MSCOCO validation set with the auxiliary supervisions from a single Swin-Small Mask RCNN \cite{he2017mask} teacher.

\subsection{Why the Auxiliary Supervisions Help?}
The auxiliary supervisions from teacher detectors lead to evident gains to the DETR-based detectors. In the following, we try to investigate its effectiveness through experimental analyses. We take $\mathcal{H}$-Deformable-DETR \cite{jia2022detrs} as the baseline with a Swin-Small backbone, 300 queries and 12-epoch training schedule. The predicted boxes from a Swin-Small Mask RCNN \cite{he2017mask} are used as auxiliary supervisions.

\vspace{-4mm}
\paragraph{Enrich supervisions.}
As shown in Fig. \ref{fig:aux_supervision}, among the teacher's boxes, there are some newly-annotated objects, \eg, the keyboard of the laptop, can be used to enrich the annotations. According to our statistics, the number of newly-annotated boxes, which have no overlap with the GT boxes, accounts for about 15\% of the total teacher's boxes.
In Tab. \ref{table:why_aux_help}, we conduct an experiment to leverage those newly-annotated boxes as auxiliary supervisions, denoted as \emph{Newly-annotated boxes}. It can be seen that, using the newly-annotated boxes can slightly improve the performance over baseline, indicating the newly-annotated boxes alone in the auxiliary supervisions can alreadly provide useful knowledge to boost student's performance.
\begin{table}[!t]
\begin{center}
\caption{Results on MSCOCO validation set for validating the effectiveness of the auxiliary supervisions. The student is a $\mathcal{H}$-Deformable-DETR \cite{jia2022detrs} with Swin-Small backbone, 300 queries and 12-epoch training schedule. Auxiliary boxes is from the Swin-small Mask-RCNN \cite{he2017mask}.}
\vspace{-2mm}
\label{table:why_aux_help}
\resizebox{1\columnwidth}{!}{
\begin{tabular}{l|cccc}
\toprule
\rowcolor{black!20}
\textbf{Method} & \textbf{AP} & \textbf{AP$_{50}$} & \textbf{AP$_{75}$} \\
\hline
\hline
Student & 52.5 & 71.1 & 57.3 \\ 
Student + Full Aux  & 53.5 & 72.0 & 58.2 \\ 
\midrule
Newly-annotated boxes & 52.7 & 71.5 & 57.9 \\ 
Replace labels with GT labels & 53.2 & 71.5 & 58.2 \\
Replace scores with IoUs & 53.3 & 71.6 & 58.5 \\
Replace labels \& scores & 53.1 & 71.3 & 58.2 \\
\bottomrule
\end{tabular}}
\vspace{-5mm}
\end{center}
\end{table}

\vspace{-4mm}
\paragraph{Teacher knowledge in box labels and confidence scores.}
From the perspective of knowledge distillation, the predicted labels of the teacher's boxes can represent the semantic relations between object categories to some extent. For the predicted boxes of Swin-Small Mask RCNN, there are about 19\% of the auxiliary boxes' labels are different from those of the GT boxes in the same image. 
Here, we replace the labels of the teacher's boxes with the labels of the GT boxes which have the highest IoUs with them. For those teacher's boxes which have no overlap with GT boxes, we leave their labels unchanged. As shown in Tab. \ref{table:why_aux_help}, after changing the predicted box labels to GT labels, the gain drops by 0.3\%, showing that the predicted labels of the teacher's boxes can help to improve DETR's performance.

Besides, the predicted confidence scores of the teacher's boxes can also encode the teacher's knowledge about the quality of the predicted boxes. Here, we replace the confidence scores with the IoUs of the teacher's boxes to their nearest GT boxes. For the teacher's boxes which have no overlap with GT boxes, we leave their confidence scores unchanged. As shown in Tab. \ref{table:why_aux_help}, after replacing the confidence scores, the gain drops slightly by 0.2\%, showing that the teacher detectors can evaluate the quality of predicted boxes more effectively, and using IoU scores is sub-optimal.

\begin{figure}[!t]
\centering
\includegraphics[width=0.95\columnwidth]{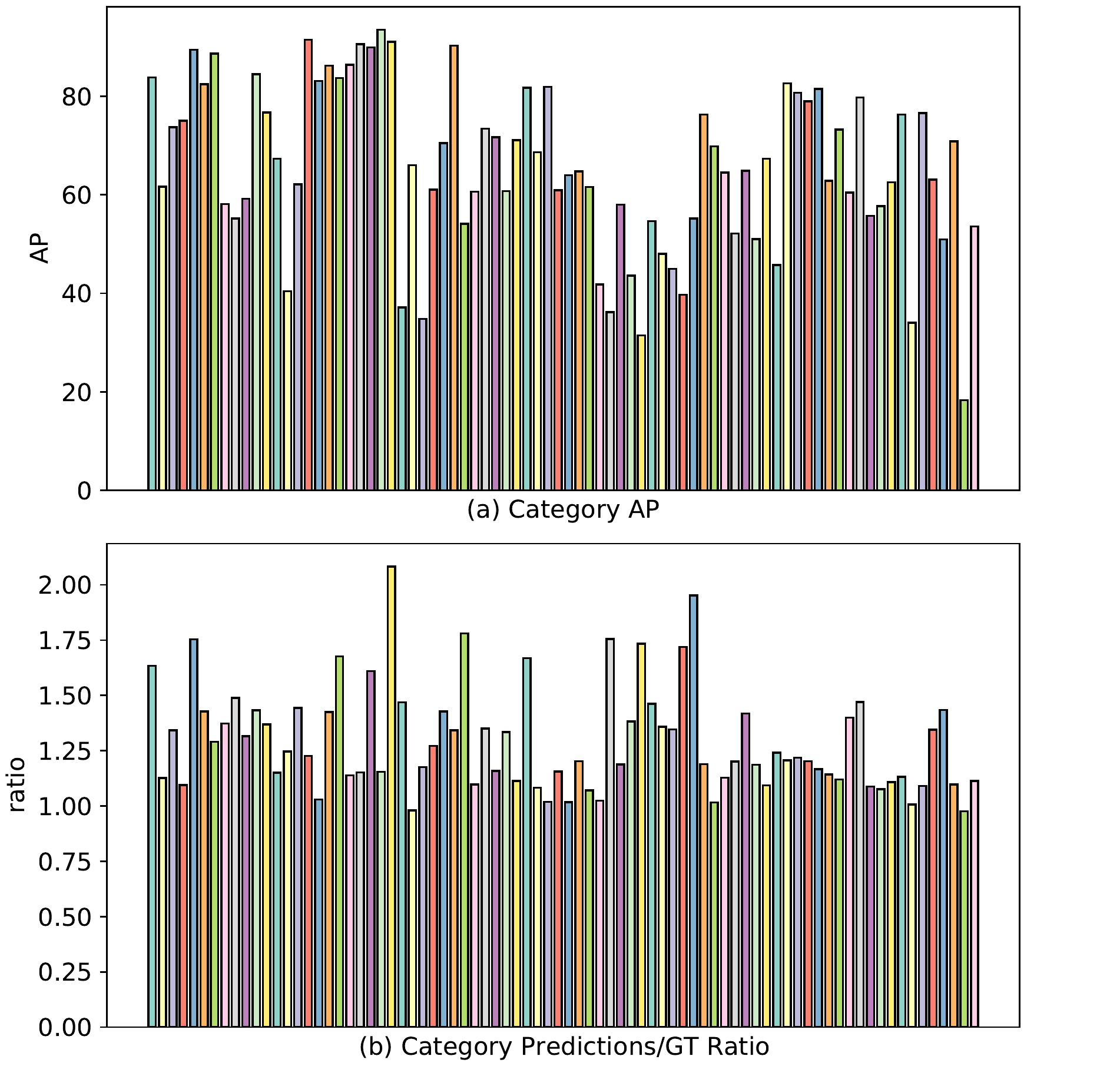}
\caption{(a) Category AP of R50 $\mathcal{H}$-Deformable-DETR \cite{jia2022detrs} on MSCOCO 2017 validation set. (b) The ratio of the number of the predicted boxes of R50 Mask RCNN \cite{he2017mask}, whose $\text{IoUs}>0.5$, to the number of GT boxes.}
\label{fig:relation_AP_number}
\vspace{-1mm}
\end{figure}

\vspace{-4mm}
\paragraph{Intrinsic characteristics of objects.}
Since teacher detectors are learned under the supervision of the GT boxes, therefore, the teacher-generated boxes would capture machine-perceived object locations and some \emph{intrinsic characteristics} of the objects, such as detection difficulties. As shown in Fig. \ref{fig:relation_AP_number}, if we take the AP value of a category to represent its detection difficulty, we can see that the lower the AP of the category, the higher the ratio of the number of auxiliary boxes to the number of GT boxes. In other words, for objects that are easy to detect, such as the ``mouse'' in Fig. \ref{fig:aux_supervision}, it is likely that each such object has only one auxiliary box, while some objects that are difficult to detect might have multiple auxiliary boxes that are more diversely distributed. In this way, during training, the student detector would pay more attention (losses) to those hard objects, which have more auxiliary boxes, and thus improve training efficiency and detection performance.

As shown in Tab. \ref{table:why_aux_help}, if we simultaneously replace the labels and the confidence scores as above, the auxiliary information from teachers can only provide the box positions. Even if, it still achieves 0.6\% gain over the baseline model, demonstrating that the positions of auxiliary boxes alone is an informative source for supervising the student DETR.

\section{Experiment} \label{sec:experiment}

\subsection{Setup}

\paragraph{Dataset.}
We demonstrate the effectiveness of Teach-DETR on the MSCOCO 2017 dataset \cite{lin2014microsoft}. Following most previous methods, we train the model on the training set and report the mean average precision (AP) on the validation set under IoU thresholds from 0.5 to 0.95 with interval 0.05. Moreover, we also optionally report the mean average precision of different object scales, including small (AP$_{S}$), medium (AP$_{M}$) and large (AP$_{L}$). To validate the effectiveness of Teach-DETR, we also report the results of compatibility of Teach-DETR on the other public dataset, LVIS v1.0 \cite{gupta2019lvis}, with mean average precision.

\vspace{-4mm}
\paragraph{Implementation details.}
Since Teach-DETR is quite efficiency and introduces no additional parameters, for fair comparison, we directly leverage the training settings of the student detectors, including batch size, learning rate, optimizer, initialization strategy, data augmentations, \etc.

The auxiliary boxes from teachers are obtained in an offline manner. Since the efficiency of bipartite matching greatly influences the efficiency of Teach-DETR, with more number of auxiliary boxes, the more time it takes to conduct bipartite matching. Therefore, in practice, we control the number of auxiliary boxes for each teacher below 50 by selecting boxes with high confidence scores. Especially for LVIS v1.0 \cite{gupta2019lvis}, the detectors are prone to generate a lot of boxes for each image, there are also too much boxes with low confidence scores (below 0.01). Even if we use the confidence scores to weight the losses of auxiliary boxes, too many low-confident boxes will still have a negative impact.

During training, we do not utilize the confidence scores of the auxiliary boxes during the bipartite matching. It is a trade-off, since according to our experiments, this way can only bring 0.1\% gain as well as more computational cost.

For $\mathcal{H}$-Deformable-DETR \cite{jia2022detrs}, we only apply the auxiliary supervisions to the one-to-one matching branch. In our experiment, additionally using the auxiliary supervisions for the one-to-many matching branch brings no further improvement.
For DN-DETR \cite{li2022dn} and DINO \cite{zhang2022dino}, we only apply the auxiliary boxes to the matching part. For the denoising part, we do not use the auxiliary supervisions.

\begin{table}[!t]
\begin{center}
\caption{Detection performance of several teacher detectors on MSCOCO 2017 validation set. Swin-Large is pretrained on the ImageNet-22k \cite{deng2009imagenet}.}
\vspace{-3mm}
\label{table:teacher_detection_results}
\resizebox{1\columnwidth}{!}{
\begin{tabular}{l|c|c|c}
\toprule
\rowcolor{black!15}
\textbf{Method} & \textbf{Backbone} & \textbf{\#epoch} & \textbf{AP} \\
\hline
\hline
\multirow{3}{*}{Mask RCNN \cite{he2017mask}}
& R50 & 36 & 40.7 \\
& Swin-S & 36 & 46.4 \\
& Swin-L & 36 & 49.4 \\
\midrule
RetinaNet \cite{lin2017focal} & Swin-S & 36 & 46.4 \\
FCOS \cite{tian2019fcos} & Swin-S & 36 & 45.4 \\
$\mathcal{H}$-Deformable-DETR \cite{jia2022detrs} & Swin-S & 12 & 52.5 \\
\bottomrule
\end{tabular}}
\vspace{-5mm}
\end{center}
\end{table}
\begin{table}[!t]
\begin{center}
\caption{Detection performance of several teacher detectors on LVIS v1.0 validation set.}
\vspace{-3mm}
\label{table:teacher_detection_results_lvis}
\begin{tabular}{l|c|c|c}
\toprule
\rowcolor{black!15}
\textbf{Method} & \textbf{Backbone} & \textbf{\#epoch} & \textbf{AP} \\
\hline
\hline
Mask RCNN \cite{he2017mask} & Swin-S & 24 & 30.4 \\
RetinaNet \cite{lin2017focal} & Swin-S & 24 & 27.9 \\
FCOS \cite{tian2019fcos} & Swin-S & 24 & 30.9 \\
\bottomrule
\end{tabular}
\vspace{-5mm}
\end{center}
\end{table}

\vspace{-4mm}
\paragraph{Teacher detectors.}
For MSCOCO, we mainly test four teacher detectors, \ie, Mask RCNN \cite{he2017mask}, RetinaNet \cite{lin2017focal}, FCOS \cite{tian2019fcos} and $\mathcal{H}$-Deformable-DETR \cite{jia2022detrs}. These detectors are designed in different manners, including the classical two-stage and ROI-based detector, one-stage detector, anchor-free detector and DETR-based detector. The detection results of these teacher detectors are shown in Tab. \ref{table:teacher_detection_results}. Besides, the performances of the three used teacher detectors in LVIS v1.0 \cite{gupta2019lvis} are shown in Tab. \ref{table:teacher_detection_results_lvis}.

\subsection{Ablation Study}
In this section, we would like to investigate different designs of utilizing auxiliary supervisions. We report results on MSCOCO 2017 validation set, and take $\mathcal{H}$-Deformable-DETR \cite{jia2022detrs} as the student with Swin-Small Backbone, 300 queries and 12-epoch training schedule. Unless explicitly stated, we utilize the auxiliary supervisions from Swin-Small Mask RCNN \cite{he2017mask} with 36-epoch training schedule. Their performances are shown in Tab. \ref{table:teacher_detection_results}. Note that, we only apply the auxiliary supervisions to the one-to-one matching branch of $\mathcal{H}$-Deformable-DETR \cite{jia2022detrs}. In our experiment, additionally using the auxiliary supervisions for the one-to-many matching branch brings no further improvement.

\paragraph{Teacher's boxes \vs Noisy boxes.}
As shown in Sec. \ref{sec:our_method}, even if we only utilize the box positions of the teacher's boxes, the improvement still exists. A natural question is that, if we manually generate some boxes as auxiliary supervisions, can the teacher-free boxes also help to improve performance? Here, we follow DN-DETR \cite{li2022dn} to add noise to GT boxes to generate some noisy boxes. Specifically, we generate 3 groups of noisy boxes with the hyper-parameters for center shifting and box scaling being set as 0.4 and 0.4, respectively.
Compared with the teacher's boxes, the noisy boxes have the same data format, \ie, the box size and location, category, and the confidence scores can be manually set as the IoUs to GT boxes, but they have no information from teachers. 
As shown in Tab. \ref{table:different_designs}, contrary to the result of using the teacher's boxes, utilizing the noisy boxes actually deteriorates the performance of the baseline model, indicating the necessity of using auxiliary boxes from teachers.

\vspace{-4mm}
\paragraph{Offline teachers \vs Online teachers.} As shown in Tab. \ref{table:different_designs}, employing online teachers achieves comparable performance with offline ones. Considering the online manner would greatly increase training memory and computational cost, we use offline teachers to first generate supervision boxes and load them from memory during training.
\begin{table}[!t]
\begin{center}
\caption{Detection results of adopting different types of auxiliary boxes in Teach-DETR. The student is Swin-Small $\mathcal{H}$-Deformable-DETR \cite{jia2022detrs}.}
\vspace{-2mm}
\label{table:different_designs}
\begin{tabular}{l|c}
\toprule
\rowcolor{black!20}
\textbf{Method} & \textbf{AP} \\
\hline
\hline
Student & 52.5 \\
Student + Offline Teacher's boxes (ours) & 53.5 \\
\midrule
Student + Noisy boxes & 51.9 \\
Student + Online Teacher's boxes & 53.6 \\
Student + Teacher's boxes w/ soft labels & 46.8 \\
\bottomrule
\end{tabular}
\vspace{-5mm}
\end{center}
\end{table}

\vspace{-4mm}
\paragraph{Hard labels \vs Soft labels.} In Teach-DETR, we use the hard labels of the predicted boxes, which are the category id with the highest probabilities. 
As shown in Tab. \ref{table:different_designs}, when we utilize the soft labels, \ie, the predicted class probabilities of boxes, the detection performance drops significantly. It is because the RCNN-based detectors, such as Mask RCNN \cite{he2017mask}, usually adopt the softmax operation and cross-entropy loss, while the DETR-based detectors prefer using sigmoid operation and binary cross-entropy loss or focal loss \cite{lin2017focal}. It is hard to find a way to align the two kinds of predictions. In contrast, the hard labels can ignore this discrepancy.
\begin{figure}[!t]
  \centering
  \includegraphics[width=1.0\columnwidth]{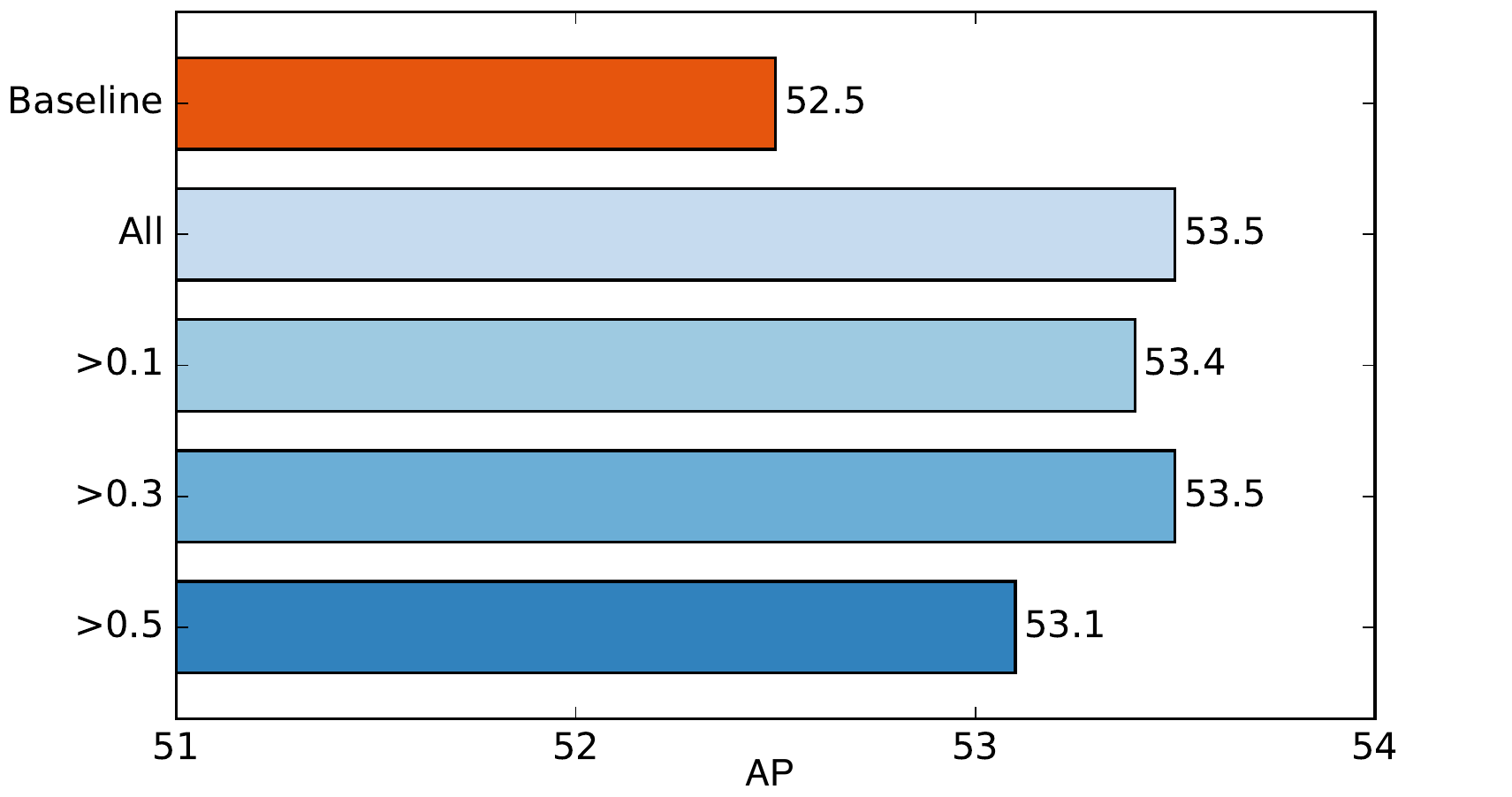}
\caption{Detection results of utilizing auxiliary boxes of different IoUs. Auxiliary boxes are from Swin-S Mask RCNN \cite{he2017mask}.}
\label{fig:investigate_aux_box_IoUs}
\end{figure}

\vspace{-4mm}
\paragraph{Utilizing auxiliary boxes of different IoUs.} 
In Sec. \ref{sec:our_method}, we have shown that utilizing the newly-annotated boxes alone can also bring performance gain over the baseline detector. Here, we step further to investigate the influences of different auxiliary boxes. We group them according to their IoUs to GT boxes. For Swin-S Mask RCNN \cite{he2017mask}, its predicted boxes, whose IoUs are $0.0$, $(0.0, 0.5]$, $(0.5, 1.0]$ to GT boxes, account for about 15\%, 51\% and 34\% of the total number of the auxiliary boxes, respectively.
To figure out which IoU's boxes contributes most to the DETR student, we divide the auxiliary boxes into three groups according to their IoUs to the GT boxes, \ie, the boxes of IoUs$>$0.1, $>$0.3 and $>$0.5, with results showing in Fig. \ref{fig:investigate_aux_box_IoUs}. As we can see, the major improvement comes from the boxes which have overlap with GT boxes. Especially, when we utilize the auxiliary boxes of IoU$>$0.3, it achieves the same performance as using the all auxiliary boxes. However, using the auxiliary boxes of IoU$>$0.1 degrades the performance, indicating that the auxiliary boxes of 0.1$<$IoU$<$0.3 may be quite noisy. When we further tighten the restriction to leverage only the auxiliary boxes of IoU$>$0.5, the performance drops evidently by 0.4\%, because of the insufficient number of boxes to stabilize training and provide complementary information from the teachers. Considering that using all auxiliary boxes also achieves the best performance and to avoid introducing extra hyper-parameters, we utilize all auxiliary boxes in our approach and other experiments.
\begin{table}[!t]
\begin{center}
    \caption{Results of applying auxiliary supervisions to different layers. The student is a $\mathcal{H}$-Deformable-DETR \cite{jia2022detrs} with Swin-Small backbone, 300 queries and 12-epoch training schedule. Auxiliary boxes is from the Swin-small Mask-RCNN \cite{he2017mask}. Enc and Dec denote the Transformer encoder and decoder, respectively.}
\vspace{-2mm}
\label{table:different_layers}
\begin{tabular}{l|cccc}
\toprule
\rowcolor{black!20}
\textbf{Method} & \textbf{AP} & \textbf{AP$_{50}$} & \textbf{AP$_{75}$} \\
\hline
\hline
Final Layer of Enc & 52.6 & 71.1 & 57.4 \\
Final Layer of Dec & 52.7 & 71.2 & 57.4 \\
Intermediate Layers of Dec & 53.3 & 72.0 & 58.3 \\
\midrule
All Layers & 53.5 & 72.0 & 58.2 \\
\bottomrule
\end{tabular}
\vspace{-5mm}
\end{center}
\end{table}

\vspace{-4mm}
\paragraph{Applying auxiliary supervisions to different layers.}
Since $\mathcal{H}$-Deformable-DETR \cite{jia2022detrs} is in a two-stage fashion, the GT boxes are used at the output of the Transformer encoder, and the output queries of the intermediate layers and the final layer of the Transformer decoder. Here, we investigate if the auxiliary boxes should be also applied to these layers. As shown in Tab. \ref{table:different_layers}, applying the auxiliary supervisions to the three types of layers would improve the performance. The main gain comes from the intermediate layers of the Transformer decoder. Considering the efficiency of employing auxiliary supervisions, we apply the auxiliary supervisions to all layers to obtain the best result.
\begin{table}[!t]
\begin{center}
\caption{Detection results with different teacher detectors or their combinations. (a) Teacher detectors with the same architecture but different backbones. (b) Different teacher detectors. (c) Teacher combinations. The student is Swin-Small $\mathcal{H}$-Deformable-DETR \cite{jia2022detrs}. The performances of teacher detectors are shown in Tab. \ref{table:teacher_detection_results}. }
\vspace{-2mm}
\label{table:different_teachers}
\resizebox{0.98\columnwidth}{!}{
\begin{tabular}{m{0.2cm}l|cccc}
\toprule
\rowcolor{black!20}
& \textbf{Teacher Detector} & \textbf{AP} & \textbf{AP$_{50}$} & \textbf{AP$_{75}$} \\
\hline
\hline
& No Teacher & 52.5 & 71.1 & 57.3 \\
\midrule
\multirow{3}{*}{(a)}
& R50 Mask RCNN & 52.9 & 71.3 & 58.1 \\
& Swin-S Mask RCNN & 53.5 & 72.0 & 58.5 \\
& Swin-L Mask RCNN & 53.4 & 71.8 & 58.4 \\
\midrule
\multirow{4}{*}{(b)}
& Swin-S RetinaNet & 53.4 & 71.9 & 58.3 \\
& Swin-S FCOS & 53.4 & 71.9 & 58.4 \\
& Swin-S $\mathcal{H}$-Deformable-DETR & 53.4 & 71.8 & 58.4 \\
& Mean Teacher (momentum 0.999) & 53.0 & 71.3 & 58.0 \\
\midrule
\multirow{5}{*}{(c)}
& Swin-S Mask RCNN &  &  &  \\
& + Swin-S RetinaNet & 53.9 & 72.4 & 59.0 \\
& + Swin-S FCOS & 54.1 & 72.7 & 59.5 \\
& + Swin-S $\mathcal{H}$-Deformable-DETR & 54.2 & 72.5 & 59.5 \\
& NMS Fusion & 53.9 & 72.5 & 59.1 \\
\bottomrule
\end{tabular}}
\vspace{-4mm}
\end{center}
\end{table}

\vspace{-4mm}
\paragraph{Detection performance of teacher detectors.} As mentioned above, using teacher detectors to promote object detection performance is not new for knowledge distillation.  
Nevertheless, the above experimental results demonstrate that our method is different from knowledge distillation, since the teacher detectors can be much inferior to the ``student'' detector.
To figure out whether better teacher leads to better DETR students, we conduct several experiments. 

In Tab. \ref{table:different_teachers}(a), we concentrate on the teacher detectors with the same architecture but different detection performances, achieved by changing their backbones.
We leverage the Mask RCNN \cite{he2017mask} of three different backbones, \ie, ResNet50, Swin-Small and Swin-Large. Due to the poorer performance of R50 Mask RCNN, even though using it as the teacher brings some performance improvement, there is still a significant performance gap compared with using Swin-Small Mask RCNN's auxiliary supervision. Moreover, adopting teacher boxes from Swin-Large Mask RCNN only obtains comparable performance to Swin-Small Mask RCNN. It suggests that the improvement of DETR performance is correlated with the teacher's performance. However, there is an upper bound to this improvement because boxes can only provide limited information from teachers.

We then evaluate the impact of teachers' architectures in Tab. \ref{table:different_teachers}(b). In addition to the Mask RCNN \cite{he2017mask}, we select the RetinaNet \cite{lin2017focal}, FCOS \cite{tian2019fcos} and $\mathcal{H}$-Deformable-DETR \cite{jia2022detrs} as teachers. We can see that the teacher detectors' architectures would not affect the performance if the detection performance of the teacher is high enough. Even if we use the teachers of the same architecture with the student, \ie, $\mathcal{H}$-Deformable-DETR, the performance improvement still exists. Based on this experiment, we further try to use the mean teacher \cite{tarvainen2017mean}, which is updated by adopting the exponential moving average with momentum of $0.999$, to provide auxiliary supervisions in an online manner. We observe 0.5\% gain under this setting, showing the effectiveness of our method and further proving that the performance improvement is less related to the teachers' architectures.

Finally, we investigate the expandability of our method in Tab. \ref{table:different_teachers}(c). On the basis of the Mask RCNN, we gradually incorporate the auxiliary supervisions of RetinaNet, FCOS and $\mathcal{H}$-Deformable-DETR. As we can see, the auxiliary supervisions from different teachers can be complementary to each other. When more teachers are introduced, the performance of the student DETR is consistently improved and finally reaches AP of 54.2\%. We also attempt to conduct ``early'' fusion (NMS Fusion in Tab. \ref{table:different_teachers}(c)) to fuse multiple teachers' boxes to form one group of auxiliary supervision. However, it is hard to set hyper-parameters of NMS that are proper for all teachers. After hyper-parameters tuning, the performance is still inferior to our solution.
\begin{table*}[!t]
\begin{center}
\caption{Results of different DETRs with proposed Teach-DETR on MSCOCO 2017 val set. The auxiliary boxes are collected from Swin-S Mask RCNN \cite{ren2015faster}, Swin-S FCOS \cite{tian2019fcos}, Swin-S RetinaNet \cite{lin2017focal} and Swin-S $\mathcal{H}$-Deformable-DETR \cite{jia2022detrs}, whose performances are shown in Tab. \ref{table:teacher_detection_results}.
All the Deformable-DETR-based methods are of two-stage. Weight decay of all $\mathcal{H}$-Deformable-DETRs is 0.0001.
$\dagger$ Tricks denotes dropout rate 0 within transformer, mixed query selection and look forward twice \cite{zhang2022dino,jia2022detrs}. $\ddag$ using top 300 predictions for evaluation.}
\vspace{-2mm}
\label{table:comparison_mscoco_val}
\resizebox{2.05\columnwidth}{!}{
\begin{tabular}{l|l|c|c|cccccc}
\toprule
\rowcolor{black!20}
\textbf{Method} & \textbf{Backbone} & \textbf{\#query} & \textbf{\#epochs} & \textbf{AP} & \textbf{AP$_{50}$} & \textbf{AP$_{75}$} & \textbf{AP$_{S}$} & \textbf{AP$_{M}$} & \textbf{AP$_{L}$}\\
\hline
\hline
Conditional-DETR-DC5 \cite{meng2021conditional} & R101 & 300 & 50 & 45.0 & 65.5 & 48.4 & 26.1 & 48.9 & 62.8 \\
Conditional-DETR-DC5 + Aux & R101 & 300 & 50 & 46.7 ($\uparrow1.7$) & 66.9 & 50.8 & 28.3 & 50.9 & 63.5 \\
\midrule
DAB-DETR-DC5 \cite{liu2022dab} & R101 & 300 & 50 & 45.8 & 65.9 & 49.3 & 27.0 & 49.8 & 63.8\\
DAB-DETR-DC5 + Aux & R101 & 300 & 50 & 48.5 ($\uparrow2.7$) & 68.1 & 52.8 & 30.5 & 52.6 & 64.4  \\
\midrule
DN-DETR-DC5 \cite{li2022dn} & R101 & 300 & 50 & 47.3 & 67.5 & 50.8 & 28.6 & 51.5 & 65.0 \\
DN-DETR-DC5 + Aux & R101 & 300 & 50 & 49.9 ($\uparrow2.6$)  & 69.5 & 54.2 & 31.7 & 53.8 & 66.7 \\
\midrule
YOLOS \cite{fang2021you} & DeiT-S \cite{touvron2021training} & 100 & 150 & 35.6 & 55.9 & 36.0 & 14.3 & 37.0 & 54.9\\
YOLOS + Aux & DeiT-S \cite{touvron2021training} & 100 & 150 & 38.0 ($\uparrow2.4$) & 58.4 & 39.5 & 17.4 & 40.1 & 56.3 \\
\midrule
ViDT \cite{song2021vidt} & Swin-S & 100 & 50 & 47.2 & 67.5 & 51.1 & 28.7 & 50.1 & 64.3 \\
ViDT + Aux & Swin-S & 100 & 50 & 49.0 ($\uparrow1.8$) & 68.7 & 53.4 & 32.1 & 52.0 & 64.9 \\
\midrule
Deformable-DETR \cite{zhu2021deformable} & Swin-S & 300 & 36 & 50.7 & 70.7 &54.8 & 32.4 & 54.3 & 67.2\\
Deformable-DETR + Aux & Swin-S & 300 & 36 & 53.2 ($\uparrow2.5$) & 72.3 & 58.3 & 37.3 & 56.9 & 68.5 \\
\midrule
Deformable-DETR \cite{zhu2021deformable} + tricks$\dagger$  & Swin-S & 300 & 36 & 53.8 & 72.8 & 58.9 & 36.5 & 57.5 & 69.0 \\
Deformable-DETR + tricks$\dagger$ + Aux & Swin-S & 300 & 36 & 55.5 ($\uparrow1.7$) & 74.2 & 61.1 & 40.1 & 59.4 & 70.5 \\
\midrule
DINO $\ddag$ \cite{zhang2022dino} & Swin-L (IN-22k, 384) & 900 & 36 & 57.8 & 76.5 & 63.2 & 40.6 & 61.8 & 73.5 \\
DINO $\ddag$ + Aux & Swin-L (IN-22k, 384) & 900 & 36 & 58.9 ($\uparrow1.1$) & 77.4 & 65.0 & 42.7 & 62.9 & 74.5 \\
\midrule
$\mathcal{H}$-Deformable-DETR \cite{jia2022detrs} & R50 & 300 & 36 & 50.0 & 68.3 & 54.4 & 32.9 & 52.7 & 65.3 \\
$\mathcal{H}$-Deformable-DETR + Aux & R50 & 300 & 36 & 51.9 ($\uparrow1.9$) & 70.1 & 57.0 & 35.2 & 54.8 & 66.3 \\
$\mathcal{H}$-Deformable-DETR \cite{jia2022detrs} & Swin-S & 300 & 36 & 54.2 & 73.0 & 59.1 & 36.8 & 57.9 & 69.6 \\
$\mathcal{H}$-Deformable-DETR + Aux & Swin-S & 300 & 36 & 55.8 ($\uparrow1.6$) & 74.3 & 61.4 & 39.0 & 59.8 & 69.9 \\
$\mathcal{H}$-Deformable-DETR \cite{jia2022detrs} & Swin-L (IN-22k) & 300 & 36 & 57.1 & 75.8 & 62.6 & 40.6 & 61.0 & 72.8\\
$\mathcal{H}$-Deformable-DETR + Aux & Swin-L (IN-22k) & 300 & 36 & 58.0 ($\uparrow0.9$) & 76.6 & 63.9 & 42.0 & 62.0 & 73.4 \\
$\mathcal{H}$-Deformable-DETR $\ddag$ \cite{jia2022detrs} & Swin-L (IN-22k) & 900 & 36 & 57.6 & 76.5 & 63.4 & 41.3 & 61.9 & 73.7 \\
$\mathcal{H}$-Deformable-DETR $\ddag$ + Aux & Swin-L (IN-22k) & 900 & 36 & 58.5 ($\uparrow0.9$) & 77.4 & 64.8 & 42.5 & 62.5 & 73.8 \\
\bottomrule
\end{tabular}}
\vspace{-4mm}
\end{center}
\end{table*}
\begin{figure}[!t]
  \centering
  \includegraphics[width=0.95\columnwidth]{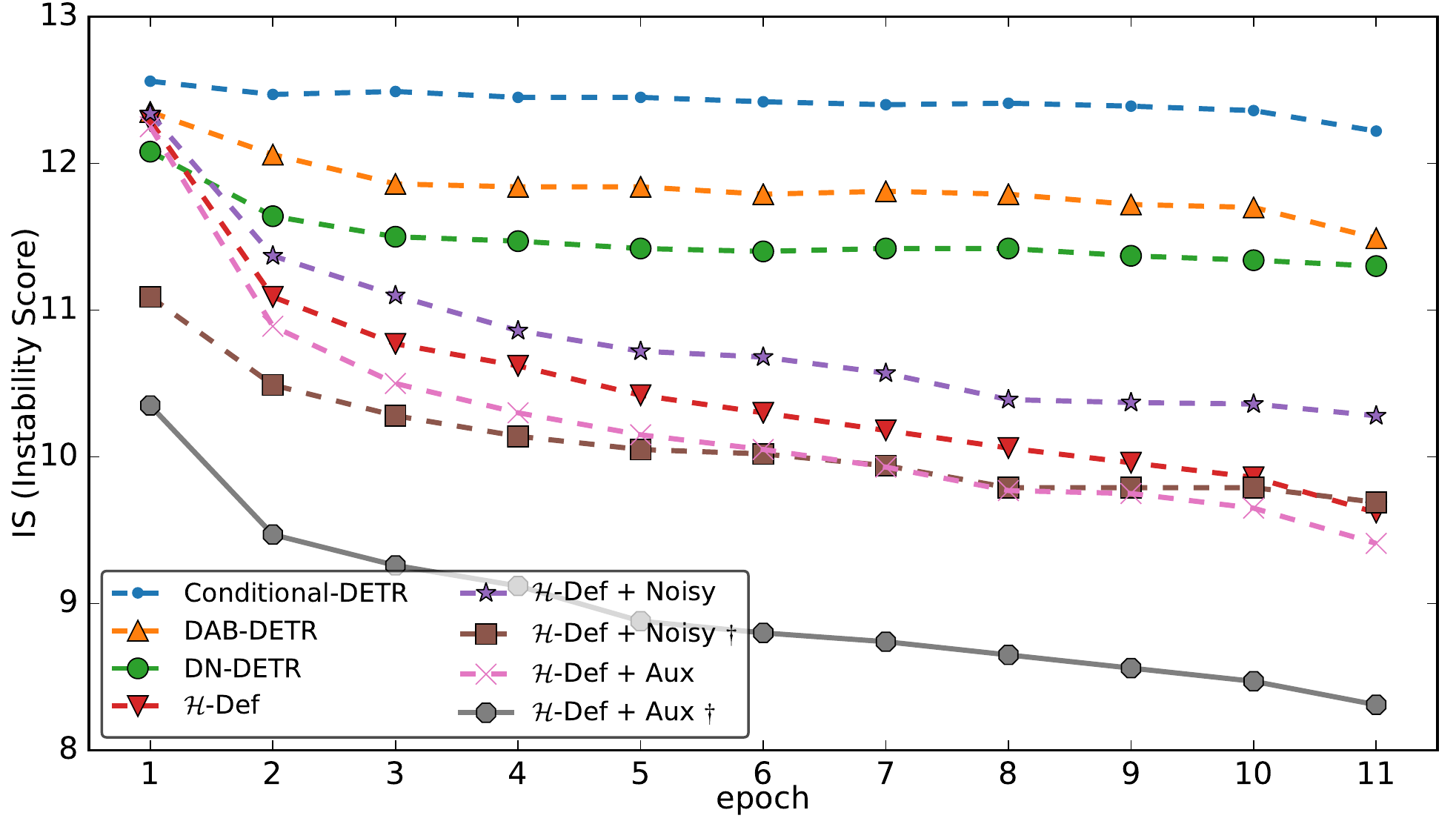}
\caption{The instability scores (IS) \cite{li2022dn} of various DETRs on MSCOCO 2017 training set. These detectors are with same settings of R50 backbone, 12-epoch training schedule and auxiliary boxes from R50 Mask RCNN (if applicable). $\dagger$ If the IoU between the auxiliary box matched by an object query and the GT box $i$ is greater than 0.5, we consider the object query is matched with the GT box $i$.}
\label{fig:instability}
\vspace{-2mm}
\end{figure}

\subsection{Analyses}
\vspace{-1mm}
\paragraph{Teach-DETR can improve stability of DETR.}
In Fig. \ref{fig:instability}, we follow DN-DETR \cite{li2022dn} to calculate the instability score (IS) of bipartite matching during training. Specifically, we store the matching results by assigning each positive query with the index of the corresponding GT box, while assigning the negative queries with index $-1$, which represents the ``no object'' class. The difference between the matching results of two adjacent epochs is viewed as the IS. It is worth noticing that after introducing the auxiliary boxes from teachers, denoted as \emph{$\mathcal{H}$-Def + Aux} in Fig. \ref{fig:instability}, the IS of $\mathcal{H}$-Deformable-DETR \cite{jia2022detrs} is reduced by $0.2$. If we take the auxiliary boxes into account, in specific, if the IoU between the auxiliary box matched by an object query and the GT box $i$ is greater than 0.5, we consider the object query is matched with the GT box $i$, the IS can be significantly reduced. In contrast, even if utilizing the noisy boxes reduces the IS at early stage, denoted as \emph{$\mathcal{H}$-Def + Noisy $\dagger$}, its final IS is even larger than the original $\mathcal{H}$-Deformable-DETR. Therefore, using auxiliary boxes from teachers can remedy the perturbation of bipartite matching to some extent, and matching either GT boxes or nearby auxiliary boxes would enforce queries to focus on the regions around objects.
\begin{figure}[!t]
  \centering
  \includegraphics[width=0.95\columnwidth]{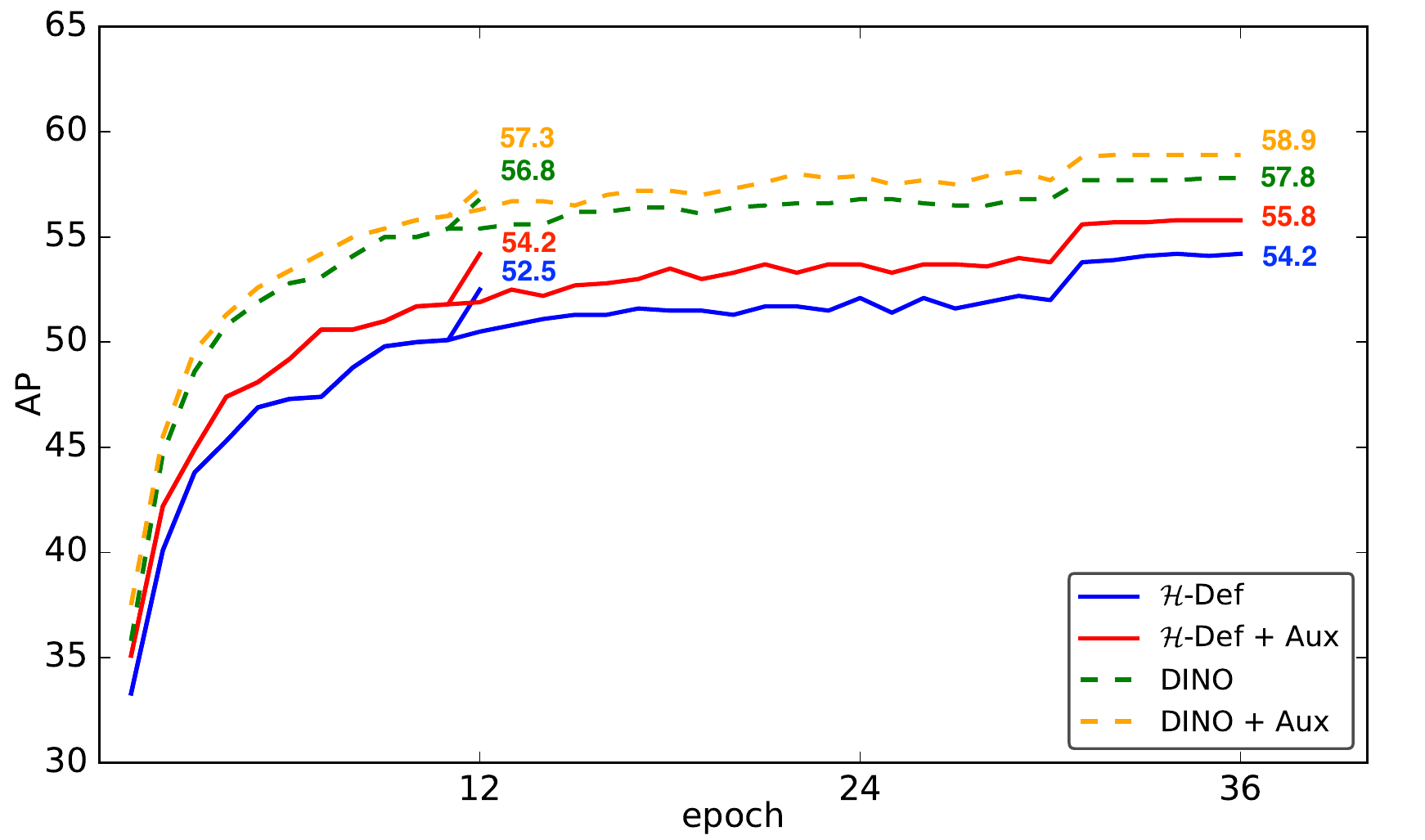}
\caption{The performance curves of Swin-Small $\mathcal{H}$-Deformable-DETR \cite{jia2022detrs}, Swin-Large DINO \cite{zhang2022dino}, and the two detectors with auxiliary supervisions. The auxiliary boxes are collected from Swin-S Mask RCNN \cite{ren2015faster}, Swin-S FCOS \cite{tian2019fcos}, Swin-S RetinaNet \cite{lin2017focal} and Swin-S $\mathcal{H}$-Deformable-DETR \cite{jia2022detrs}.}
\label{fig:performance_curve}
\end{figure}

\vspace{-4mm}
\paragraph{Performance curve.}
In Fig. \ref{fig:performance_curve}, we show performance curves of Swin-Small $\mathcal{H}$-Deformable-DETR \cite{jia2022detrs}, Swin-Large DINO \cite{zhang2022dino} and the two detectors with the auxiliary boxes from four teachers
Since Teach-DETR introduces negligible training time to the student DETR, the performances of the two detectors can be fairly compared at the same epoch. We can see that, Teach-DETR can boost the performance at the beginning and achieves evident gain at each training epoch.

\vspace{-4mm}
\paragraph{Training memory and training time.}
Compared with the baseline models, our method only introduces negligible computational cost. First, the bipartite matching can be conducted on CPU, it would not increase the GPU memory usage. For Swin-Small $\mathcal{H}$-Deformable-DETR with the auxiliary boxes from four teachers, Teach-DETR only increases the GPU memory of one GPU from 32.1G to 32.2G.
Beside, thanks to the fast computation of bipartite matching and small number of auxiliary boxes, our method only increases the training time for one epoch from 93 minutes to 94$\sim$97 minutes. The training times and memory are measured with 8 Tesla A100 GPUs.
\begin{table}[!t]
\begin{center}
\caption{Results of different DETRs with proposed Teach-DETR on LVIS v1.0 \cite{gupta2019lvis} validation set. The auxiliary supervisions are collected from Swin-S Mask RCNN \cite{ren2015faster}, Swin-S FCOS \cite{tian2019fcos}, Swin-S RetinaNet \cite{lin2017focal}, whose performances are shown in the supplementary material. Both of the two detectors use 300 queries.}
\vspace{-2mm}
\label{table:comparison_lvis_val}
\resizebox{1\columnwidth}{!}{
\begin{tabular}{l|l|c|c}
\toprule
\rowcolor{black!20}
\textbf{Method} & \textbf{Backbone} & \textbf{\#epochs} & \textbf{AP}\\
\hline
\hline
DAB-DETR-DC5 \cite{liu2022dab} & R101 & 50 & 23.7 \\
DAB-DETR-DC5 + Aux & R101 & 50 & 25.6 \\
\midrule
$\mathcal{H}$-Deformable-DETR \cite{jia2022detrs} & Swin-S & 36 & 35.7 \\
$\mathcal{H}$-Deformable-DETR \cite{jia2022detrs} + Aux & Swin-S & 36 & 37.0 \\
\bottomrule
\end{tabular}}
\vspace{-5mm}
\end{center}
\end{table}

\subsection{Compatibility with State-of-the-art DETRs}
We first verify the compatibility of Teach-DETR on MSCOCO 2017 \cite{lin2014microsoft} based on several typical DETR-based detectors. We report results on three DETR-based detectors, including Conditional DETR \cite{meng2021conditional}, DAB-DETR \cite{liu2022dab}, DN-DETR \cite{li2022dn}, and four Deformable-DETR-based methods with two-stage setting, including the vanilla Deformable-DETR \cite{zhu2021deformable}, Deformable-DETR + tricks \cite{zhang2022dino,jia2022detrs}, DINO \cite{zhang2022dino} and $\mathcal{H}$-Deformable-DETR \cite{jia2022detrs}. We report the results on their best-performance model. For $\mathcal{H}$-Deformable-DETR \cite{jia2022detrs}, we report the results with different backbones and query numbers. As in Tab. \ref{table:comparison_mscoco_val}, Teach-DETR achieves consistent improvement over baseline detectors. Specifically, it achieves 2.6\% gain on DN-DETR \cite{li2022dn}, which additionally introduces a box denoising task. For DINO \cite{zhang2022dino}, one of the top-performance DETR-based detectors, Teach-DETR can also improve its 4-scale model by 1.1\%. For $\mathcal{H}$-Deformable-DETR \cite{jia2022detrs}, which utilizes the hybrid matching, our method further boost its best model by 0.9\%.

Besides, we further verify the compatibility based on two atypical DETR-based detectors, \ie, YOLOS \cite{fang2021you} and ViDT \cite{song2021vidt}. 
YOLOS \cite{fang2021you} is an extension of ViT to object detection. It combines the backbone and the Transformer encoder together to form an efficient architecture. Besides, its decoder is a lightweight architecture, therefore, we can thus view it as an encoder-only DETR-based detector.
Likewise, ViDT \cite{song2021vidt} also integrates the backbone and the Transformer encoder together, but it keeps the Transformer decoder. We can denote it as an encoder-free DETR-based detector. We can see that, Teach-DETR can consistently boost the performances of the two DETR-based detectors, even if they are not in a typical DETR-style, demonstrating the universality of the proposed Teach-DETR.

In Tab. \ref{table:comparison_lvis_val}, we can see that Teach-DETR can also achieve evident gains over the baseline detectors on LVIS v1.0 \cite{gupta2019lvis}.

\section{Conclusion}
This paper presents a novel Teach-DETR to learn better DETR-based detectors from teacher detectors.
We show that the predicted boxes from teacher detectors are effective medium to transfer knowledge of teacher detectors to train a more accurate and robust DETR model. This new training scheme can easily incorporate the predicted boxes from multiple teacher detectors, each of which provide parallel supervisions to the student DETR. The matchings are also properly weighted according to the teachers’ confidences on the predicted boxes. During training, Teach-DETR introduces negligible computational cost, memory usage and has no requirement on teacher architectures, which is more general to various types of teacher detectors. Extensive experiments show that Teach-DETR can lead to consistent improvement for various DETR-based detectors.

{\small
\bibliographystyle{ieee_fullname}
\bibliography{egbib}

\begin{thebibliography}{10}\itemsep=-1pt

\bibitem{carion2020end}
Nicolas Carion, Francisco Massa, Gabriel Synnaeve, Nicolas Usunier, Alexander
  Kirillov, and Sergey Zagoruyko.
\newblock End-to-end object detection with transformers.
\newblock In {\em ECCV}, pages 213--229. Springer, 2020.

\bibitem{chen2017learning}
Guobin Chen, Wongun Choi, Xiang Yu, Tony Han, and Manmohan Chandraker.
\newblock Learning efficient object detection models with knowledge
  distillation.
\newblock {\em NeurIPS}, 30, 2017.

\bibitem{chen2022group}
Qiang Chen, Xiaokang Chen, Gang Zeng, and Jingdong Wang.
\newblock Group detr: Fast training convergence with decoupled one-to-many
  label assignment.
\newblock {\em arXiv preprint arXiv:2207.13085}, 2022.

\bibitem{dai2021general}
Xing Dai, Zeren Jiang, Zhao Wu, Yiping Bao, Zhicheng Wang, Si Liu, and Erjin
  Zhou.
\newblock General instance distillation for object detection.
\newblock In {\em CVPR}, pages 7842--7851, 2021.

\bibitem{deng2009imagenet}
Jia Deng, Wei Dong, Richard Socher, Li-Jia Li, Kai Li, and Li Fei-Fei.
\newblock Imagenet: A large-scale hierarchical image database.
\newblock In {\em CVPR}, pages 248--255. Ieee, 2009.

\bibitem{fang2021you}
Yuxin Fang, Bencheng Liao, Xinggang Wang, Jiemin Fang, Jiyang Qi, Rui Wu,
  Jianwei Niu, and Wenyu Liu.
\newblock You only look at one sequence: Rethinking transformer in vision
  through object detection.
\newblock {\em NeurIPS}, 34:26183--26197, 2021.

\bibitem{gao2021fast}
Peng Gao, Minghang Zheng, Xiaogang Wang, Jifeng Dai, and Hongsheng Li.
\newblock Fast convergence of detr with spatially modulated co-attention.
\newblock In {\em ICCV}, pages 3621--3630, 2021.

\bibitem{guo2021distilling}
Jianyuan Guo, Kai Han, Yunhe Wang, Han Wu, Xinghao Chen, Chunjing Xu, and Chang
  Xu.
\newblock Distilling object detectors via decoupled features.
\newblock In {\em CVPR}, pages 2154--2164, 2021.

\bibitem{gupta2019lvis}
Agrim Gupta, Piotr Dollar, and Ross Girshick.
\newblock {LVIS}: A dataset for large vocabulary instance segmentation.
\newblock In {\em CVPR}, 2019.

\bibitem{he2017mask}
Kaiming He, Georgia Gkioxari, Piotr Doll{\'a}r, and Ross Girshick.
\newblock Mask r-cnn.
\newblock In {\em ICCV}, pages 2961--2969, 2017.

\bibitem{he2016deep}
Kaiming He, Xiangyu Zhang, Shaoqing Ren, and Jian Sun.
\newblock Deep residual learning for image recognition.
\newblock In {\em CVPR}, pages 770--778, 2016.

\bibitem{hinton2015distilling}
Geoffrey Hinton, Oriol Vinyals, Jeff Dean, et~al.
\newblock Distilling the knowledge in a neural network.
\newblock {\em arXiv preprint arXiv:1503.02531}, 2(7), 2015.

\bibitem{jia2022detrs}
Ding Jia, Yuhui Yuan, Haodi He, Xiaopei Wu, Haojun Yu, Weihong Lin, Lei Sun,
  Chao Zhang, and Han Hu.
\newblock Detrs with hybrid matching.
\newblock {\em arXiv preprint arXiv:2207.13080}, 2022.

\bibitem{kuhn1955hungarian}
Harold~W Kuhn.
\newblock The hungarian method for the assignment problem.
\newblock {\em Naval Research Logistics Quarterly}, 2(1-2):83--97, 1955.

\bibitem{li2022dn}
Feng Li, Hao Zhang, Shilong Liu, Jian Guo, Lionel~M Ni, and Lei Zhang.
\newblock Dn-detr: Accelerate detr training by introducing query denoising.
\newblock In {\em CVPR}, pages 13619--13627, 2022.

\bibitem{li2022knowledge}
Gang Li, Xiang Li, Yujie Wang, Shanshan Zhang, Yichao Wu, and Ding Liang.
\newblock Knowledge distillation for object detection via rank mimicking and
  prediction-guided feature imitation.
\newblock In {\em AAAI}, volume~36, pages 1306--1313, 2022.

\bibitem{li2017mimicking}
Quanquan Li, Shengying Jin, and Junjie Yan.
\newblock Mimicking very efficient network for object detection.
\newblock In {\em CVPR}, pages 6356--6364, 2017.

\bibitem{lin2017focal}
Tsung-Yi Lin, Priya Goyal, Ross Girshick, Kaiming He, and Piotr Doll{\'a}r.
\newblock Focal loss for dense object detection.
\newblock In {\em ICCV}, pages 2980--2988, 2017.

\bibitem{lin2014microsoft}
Tsung-Yi Lin, Michael Maire, Serge Belongie, James Hays, Pietro Perona, Deva
  Ramanan, Piotr Doll{\'a}r, and C~Lawrence Zitnick.
\newblock Microsoft coco: Common objects in context.
\newblock In {\em ECCV}, pages 740--755. Springer, 2014.

\bibitem{liu2022dab}
Shilong Liu, Feng Li, Hao Zhang, Xiao Yang, Xianbiao Qi, Hang Su, Jun Zhu, and
  Lei Zhang.
\newblock Dab-detr: Dynamic anchor boxes are better queries for detr.
\newblock In {\em ICLR}, 2022.

\bibitem{liu2021swin}
Ze Liu, Yutong Lin, Yue Cao, Han Hu, Yixuan Wei, Zheng Zhang, Stephen Lin, and
  Baining Guo.
\newblock Swin transformer: Hierarchical vision transformer using shifted
  windows.
\newblock In {\em ICCV}, pages 10012--10022, 2021.

\bibitem{meng2021conditional}
Depu Meng, Xiaokang Chen, Zejia Fan, Gang Zeng, Houqiang Li, Yuhui Yuan, Lei
  Sun, and Jingdong Wang.
\newblock Conditional detr for fast training convergence.
\newblock In {\em ICCV}, pages 3651--3660, 2021.

\bibitem{redmon2016you}
Joseph Redmon, Santosh Divvala, Ross Girshick, and Ali Farhadi.
\newblock You only look once: Unified, real-time object detection.
\newblock In {\em CVPR}, pages 779--788, 2016.

\bibitem{ren2015faster}
Shaoqing Ren, Kaiming He, Ross Girshick, and Jian Sun.
\newblock Faster r-cnn: Towards real-time object detection with region proposal
  networks.
\newblock {\em NeurIPS}, 28, 2015.

\bibitem{song2021vidt}
Hwanjun Song, Deqing Sun, Sanghyuk Chun, Varun Jampani, Dongyoon Han, Byeongho
  Heo, Wonjae Kim, and Ming-Hsuan Yang.
\newblock Vidt: An efficient and effective fully transformer-based object
  detector.
\newblock In {\em ICLR}, 2021.

\bibitem{sun2020distilling}
Ruoyu Sun, Fuhui Tang, Xiaopeng Zhang, Hongkai Xiong, and Qi Tian.
\newblock Distilling object detectors with task adaptive regularization.
\newblock {\em arXiv preprint arXiv:2006.13108}, 2020.

\bibitem{tarvainen2017mean}
Antti Tarvainen and Harri Valpola.
\newblock Mean teachers are better role models: Weight-averaged consistency
  targets improve semi-supervised deep learning results.
\newblock {\em NeurIPS}, 30, 2017.

\bibitem{tian2019fcos}
Zhi Tian, Chunhua Shen, Hao Chen, and Tong He.
\newblock Fcos: Fully convolutional one-stage object detection.
\newblock In {\em ICCV}, pages 9627--9636, 2019.

\bibitem{touvron2021training}
Hugo Touvron, Matthieu Cord, Matthijs Douze, Francisco Massa, Alexandre
  Sablayrolles, and Herv{\'e} J{\'e}gou.
\newblock Training data-efficient image transformers \& distillation through
  attention.
\newblock pages 10347--10357. PMLR, 2021.

\bibitem{vaswani2017attention}
Ashish Vaswani, Noam Shazeer, Niki Parmar, Jakob Uszkoreit, Llion Jones,
  Aidan~N Gomez, {\L}ukasz Kaiser, and Illia Polosukhin.
\newblock Attention is all you need.
\newblock {\em NeurIPS}, 30, 2017.

\bibitem{wang2019distilling}
Tao Wang, Li Yuan, Xiaopeng Zhang, and Jiashi Feng.
\newblock Distilling object detectors with fine-grained feature imitation.
\newblock In {\em CVPR}, pages 4933--4942, 2019.

\bibitem{wang2022anchor}
Yingming Wang, Xiangyu Zhang, Tong Yang, and Jian Sun.
\newblock Anchor detr: Query design for transformer-based detector.
\newblock In {\em AAAI}, volume~36, pages 2567--2575, 2022.

\bibitem{yang2022focal}
Zhendong Yang, Zhe Li, Xiaohu Jiang, Yuan Gong, Zehuan Yuan, Danpei Zhao, and
  Chun Yuan.
\newblock Focal and global knowledge distillation for detectors.
\newblock In {\em Proceedings of the IEEE/CVF Conference on Computer Vision and
  Pattern Recognition}, pages 4643--4652, 2022.

\bibitem{yao2021g}
Lewei Yao, Renjie Pi, Hang Xu, Wei Zhang, Zhenguo Li, and Tong Zhang.
\newblock G-detkd: Towards general distillation framework for object detectors
  via contrastive and semantic-guided feature imitation.
\newblock In {\em ICCV}, pages 3591--3600, 2021.

\bibitem{zhang2022dino}
Hao Zhang, Feng Li, Shilong Liu, Lei Zhang, Hang Su, Jun Zhu, Lionel~M Ni, and
  Heung-Yeung Shum.
\newblock Dino: Detr with improved denoising anchor boxes for end-to-end object
  detection.
\newblock {\em arXiv preprint arXiv:2203.03605}, 2022.

\bibitem{zhang2020improve}
Linfeng Zhang and Kaisheng Ma.
\newblock Improve object detection with feature-based knowledge distillation:
  Towards accurate and efficient detectors.
\newblock In {\em ICLR}, 2020.

\bibitem{zheng2022localization}
Zhaohui Zheng, Rongguang Ye, Ping Wang, Dongwei Ren, Wangmeng Zuo, Qibin Hou,
  and Ming-Ming Cheng.
\newblock Localization distillation for dense object detection.
\newblock In {\em CVPR}, pages 9407--9416, 2022.

\bibitem{zhixing2021distilling}
Du Zhixing, Rui Zhang, Ming Chang, Shaoli Liu, Tianshi Chen, Yunji Chen, et~al.
\newblock Distilling object detectors with feature richness.
\newblock {\em NeurIPS}, 34:5213--5224, 2021.

\bibitem{zhu2021deformable}
Xizhou Zhu, Weijie Su, Lewei Lu, Bin Li, Xiaogang Wang, and Jifeng Dai.
\newblock Deformable detr: Deformable transformers for end-to-end object
  detection.
\newblock In {\em ICLR}, 2021.

\end{thebibliography}
}

\end{document}